\documentclass[letterpaper]{article} % DO NOT CHANGE THIS
\usepackage{aaai24}  % DO NOT CHANGE THIS
\usepackage{times}  % DO NOT CHANGE THIS
\usepackage{helvet}  % DO NOT CHANGE THIS
\usepackage{courier}  % DO NOT CHANGE THIS
\usepackage[hyphens]{url}  % DO NOT CHANGE THIS
\usepackage{graphicx} % DO NOT CHANGE THIS
\usepackage{natbib}  % DO NOT CHANGE THIS AND DO NOT ADD ANY OPTIONS TO IT
\usepackage{caption} % DO NOT CHANGE THIS AND DO NOT ADD ANY OPTIONS TO IT
\usepackage{algorithm}
\usepackage{algorithmic}

\usepackage{microtype}
\usepackage{subcaption}
\usepackage{booktabs} % for professional tables
\usepackage{multirow}
\usepackage{colortbl}  
\usepackage[svgnames]{xcolor}
\usepackage{array}  
\usepackage{amssymb}

\usepackage{newfloat}
\usepackage{listings}
\DeclareCaptionStyle{ruled}{labelfont=normalfont,labelsep=colon,strut=off} % DO NOT CHANGE THIS
\floatstyle{ruled}
\newfloat{listing}{tb}{lst}{}
\floatname{listing}{Listing}
\title{Transformer as Linear Expansion of Learngene}
\author{
    Shi-Yu Xia, Miaosen Zhang, Xu Yang\footnotemark[1], Ruiming Chen, Haokun Chen, Xin Geng\thanks{Co-corresponding author.}\\
}
\affiliations{
    School of Computer Science and Engineering, Southeast University, Nanjing 210096, China\\
    Key Laboratory of New Generation Artificial Intelligence Technology and Its Interdisciplinary Applications (Southeast University), Ministry of Education, China\\
    \{shiyu\_xia, 230228501, 101013120, 213193308, chenhaokun, xgeng\}@seu.edu.cn
}

\usepackage{bibentry}
\begin{document}

\maketitle

%%%%%%%%%%%%%%%%%%%%%%%%%%% ABSTRACT
\begin{abstract}
We propose expanding the shared Transformer module to produce and initialize Transformers of varying depths, enabling adaptation to diverse resource constraints.
Drawing an analogy to genetic expansibility, we term such module as \textit{learngene}.
To identify the expansion mechanism, we delve into the relationship between the layer's position and its corresponding weight value, and find that \textit{linear function} appropriately approximates this relationship.
Building on this insight, we present \textbf{T}ransformer as \textbf{L}inear \textbf{E}xpansion of learn\textbf{G}ene~(\textbf{TLEG}), a novel approach for flexibly producing and initializing Transformers of diverse depths.
Specifically, to learn learngene, we firstly construct an auxiliary Transformer linearly expanded from learngene, after which we train it through employing soft distillation.
Subsequently, we can produce and initialize Transformers of varying depths via linearly expanding the well-trained learngene, thereby supporting diverse downstream scenarios.
Extensive experiments on ImageNet-1K demonstrate that TLEG achieves comparable or better performance in contrast to many individual models trained from scratch, while reducing around 2$\times$ training cost.
When transferring to several downstream classification datasets, TLEG surpasses existing initialization methods by a large margin (\textit{e}.\textit{g}., +6.87\% on iNat 2019 and +7.66\% on CIFAR-100).
Under the situation where we need to produce models of varying depths adapting for different resource constraints, TLEG achieves comparable results while reducing around 19$\times$ parameters stored to initialize these models and around 5$\times$ pre-training costs, in contrast to the pre-training and fine-tuning approach.
When transferring a fixed set of parameters to initialize different models, TLEG presents better flexibility and competitive performance while reducing around 2.9$\times$ parameters stored to initialize, compared to the pre-training approach.
\end{abstract}

%%%%%%%%%%%%%%%%%%%%%%%%%%% Introduction
\section{Introduction}
\label{intro}
Deep neural networks (DNNs), \textit{e}.\textit{g}., Vision Transformer, have demonstrated remarkable performance in a wide variety of computer vision tasks~\cite{sun2019meta, carion2020end, liang2020polytransform, dosovitskiy2020image, zhang2021few, qin2023transferability}.
Parameter initialization is a pivotal step prior to training and wields a critical influence over the ultimate quality of the trained network~\cite{glorot2010understanding, he2016deep, arpit2019initialize, huang2020improving, zhang2021self, czyzewski2022breaking}.
Nowadays, large-scale pre-training on massive curated data brings huge \emph{foundation models}, which furnishes a superb starting point for fine-tuning across diverse downstream tasks~\cite{liu2021swin, oquab2023dinov2}.
However, the parameters of original whole model are required storing and updating separately for each downstream task during the popular pre-training and fine-tuning process, which is prohibitively expensive and time-consuming for the current ever-increasing capacity of vision models.
Furthermore, this approach lacks the flexibility to initialize models of \emph{varying scales} to meet diverse scenario demands, such as edge and IoT devices with constrained computational resources.
Therefore, in different application scenarios, a fundamental research question naturally arises: \emph{how to efficiently produce and initialize individual models considering both the model performance and resource constraint?}

\begin{figure}[t]
    \centering
    \begin{subfigure}{0.49\linewidth}
        \includegraphics[scale=0.155]{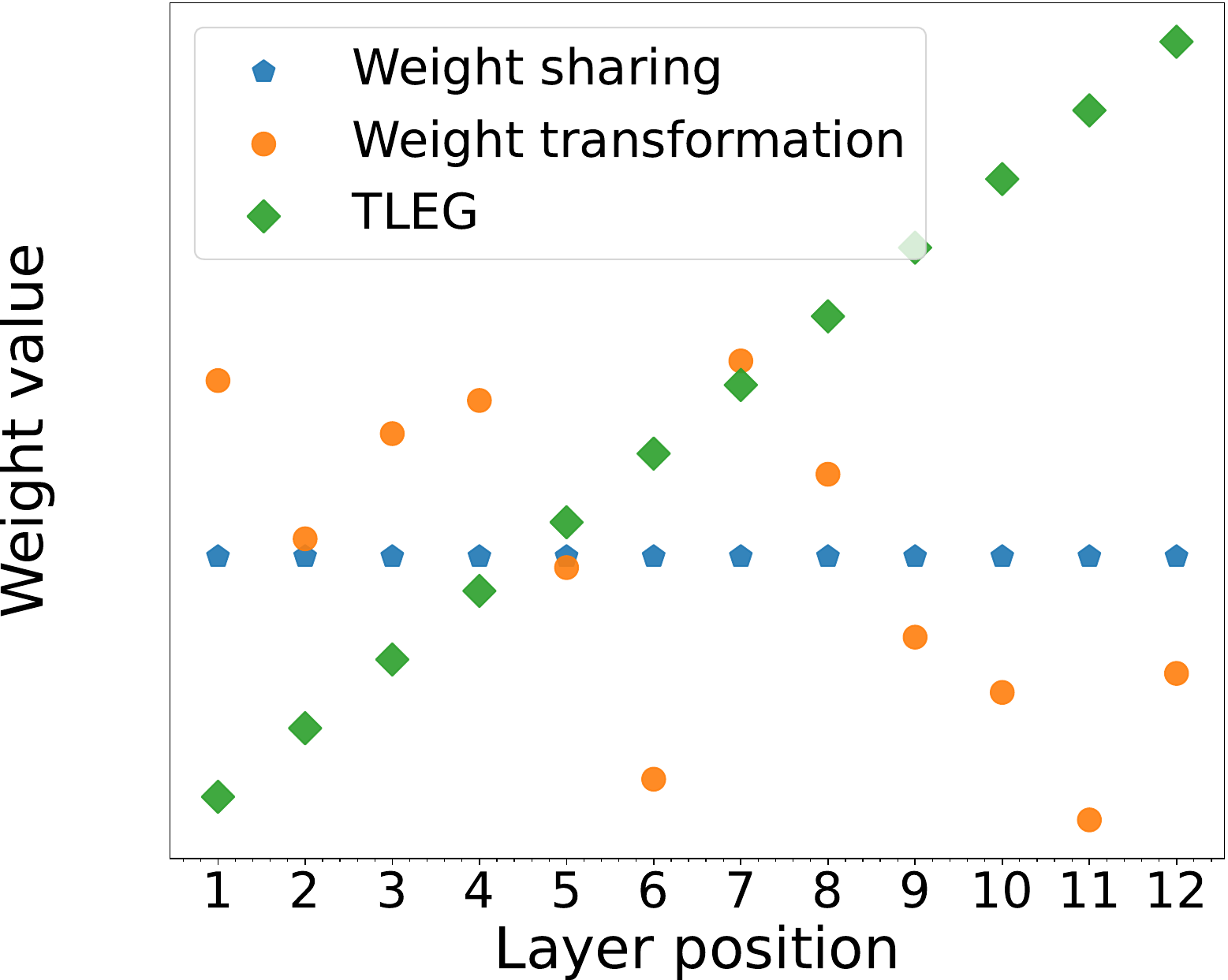}
        % \caption{}
        % \label{fig:intro1-c}
    \end{subfigure}
    \centering
    \begin{subfigure}{0.49\linewidth}
        \includegraphics[scale=0.155]{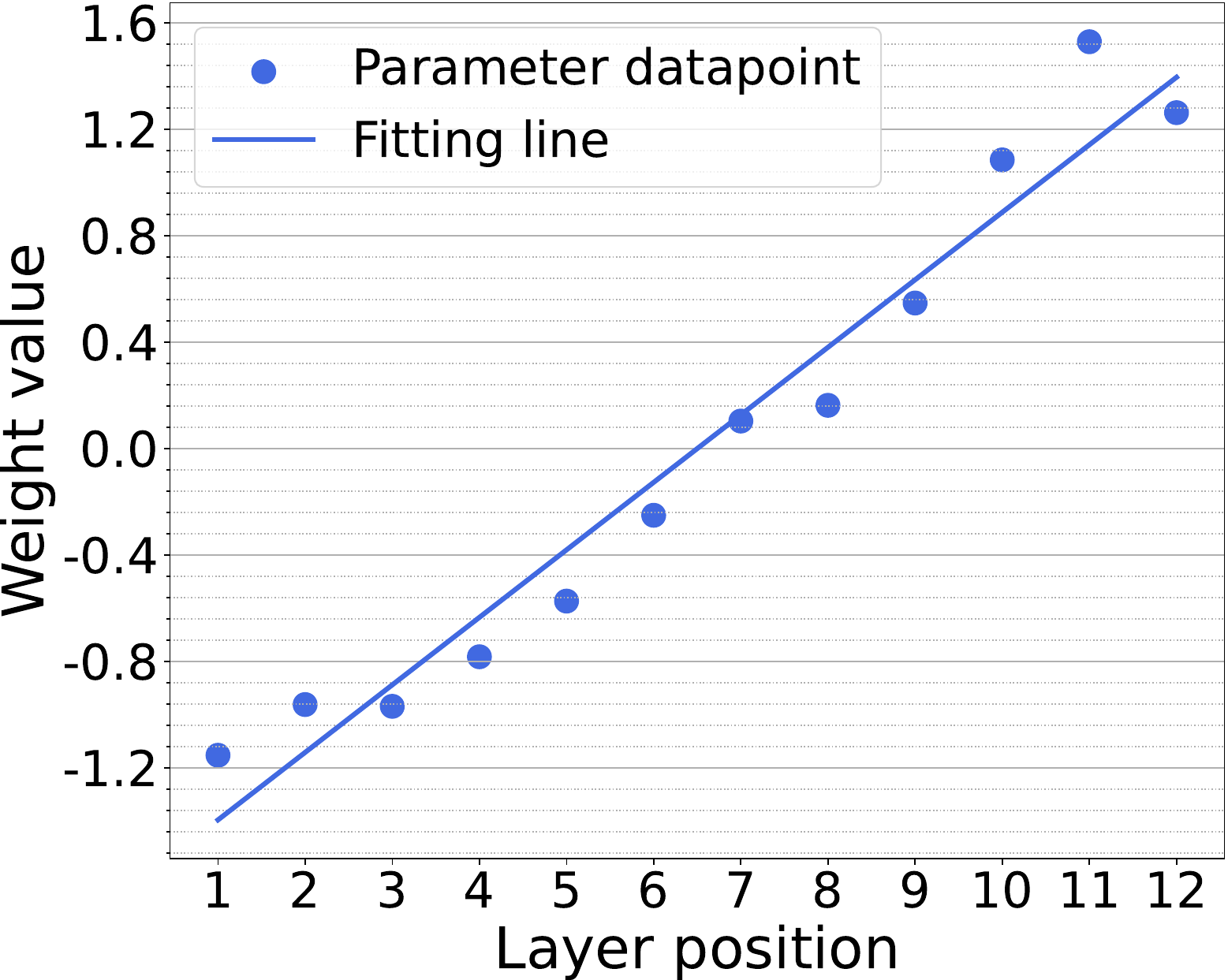}
        % \caption{}
        % \label{fig:intro1-d}
    \end{subfigure}
    \caption{
    Left: the relationship between the layer's position and its corresponding weight value of different methods. Right: our empirical observation of such relationship based on ViT-B, which shows approximate linear relationship.
    }
    \label{fig:intro1}
\end{figure}

\begin{figure}[t]
    \centering
    \includegraphics[scale=0.38]{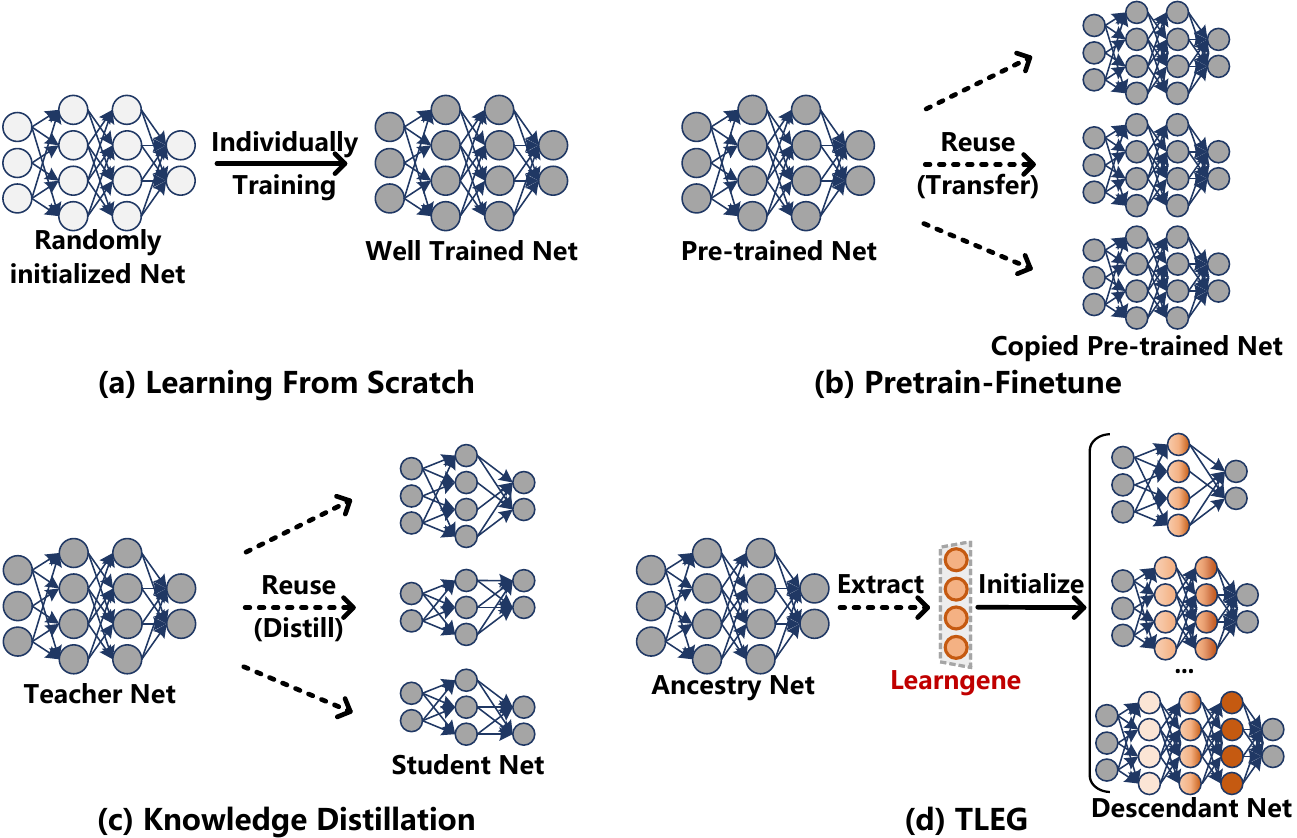}
    \caption{
    (a) Learning from scratch randomly initializes different networks in varied applications, where the training and storage costs increase linearly with the number of possible cases.
    (b) Pretrain-Finetune stores and reuses the original whole model every time facing different scenarios.
    (c) Distillation transfers knowledge from a large teacher net to a smaller student one, which requires forward propagation on teacher every time training new students.
    (d) Our TLEG directly extracts one compact learngene from an ancestry net \emph{once} and quickly initializes new descendant nets with our linear expansion strategy, allowing adaptation to diverse resource constraints.
    Note that each dotted arrow means that we need to reuse the ancestry/pretrained/teacher net once.
    }
    \label{fig:intro2}
\end{figure}

Mimicking the behaviour of the organismal gene,~\cite{wang2022learngene, wang2023learngene} proposed an innovative learning framework known as \emph{Learngene} which firstly learns the condensed knowledge termed as \textbf{learngene} from the ancestry model, and then inherits this small part to initialize descendant/downstream models.
The existing work He-LG~\cite{wang2022learngene} extracts a few integral layers as learngene based on the gradient information of the ancestry model, after which the descendant models are constructed by stacking the randomly initialized low-level layers with the extracted learngene layers.
Nevertheless, there are three major limitations existed in~\cite{wang2022learngene}.
Firstly, the strategies of extracting and utilizing the learngene are inconsistent, yielding diminished performance.
Secondly, He-LG ignores descendant models across different scales.
Lastly, He-LG does not explore Transformer-based architectures, with which the performance is also unsatisfactory.

As mentioned before, Learngene is dedicated to retaining the most generalizable part of the ancestry model, which naturally directs our attention towards eliminating redundant parameters.
% As mentioned before, Learngene is committed to retaining the most generalizable part of the ancestry model, which naturally motivates us to firstly consider eliminating redundant parameters.
One prominent approach that exemplifies this endeavor is weight sharing~\cite{lan2019albert, zhang2022minivit}, which shares identical parameters across all layers to maximise parameter elimination.
% As one of the most representative work, weight sharing~\cite{lan2019albert, zhang2022minivit} directly shares all the layers with the same parameters to maximise the elimination.
Despite its simplicity, such fully-sharing method notably compromises the model capabilities~\cite{zhang2022minivit}.
To alleviate this problem, researchers~\cite{zhang2022minivit} apply weight transformation, which imposes learnable functions on the shared weights to increase parameter diversity.
Interestingly, if we treat the parameters of each layer as one high-dimensional tensor, we can illustrate the relationship between the layer's position and its corresponding parameter value, as shown in the left portion of Fig.~\ref{fig:intro1}.
% Interestingly, if we treat the parameters of each layer as one high-dimensional tensor, we can sketch the relationship between the layer position and corresponding parameter value, as shown in Fig.~\ref{fig:intro1-c}.
Specifically, weight sharing presents a ``horizontal line'' as each layer shares identical parameters.
Correspondingly, weight transformation~\cite{zhang2022minivit} scatters the parameters due to the layer-specific mapping function.
% Specifically, weight sharing method presents a ``horizontal line'' function that each layer uses the same parameters, while weight transformation method~\cite{zhang2022minivit} scatters the parameters due to the layer-wise mapping function.
Upon closer observation, we wonder if there exists a intermediate situation between them, \textit{i}.\textit{e}., is there any simpler function that could approximate the relationship between the layer's position and its corresponding parameter value?
% Observing them, we may wonder if there exists a middle situation between them, \ie, is there any simpler function that could approximate the relationship between layer position and corresponding parameter value, and meantime does not introduce additional learnable functions?

To obtain some empirical observations of such relationship, we use PCA~\cite{karamizadeh2013overview} to transform each tensor to 1-D data point for convenience.
% To get some empirical observations, we use PCA~\cite{karamizadeh2013overview} to reduce the dimension of each tensor into 1-D data point for convenience.
Here we choose the well-trained ViT-B~\cite{dosovitskiy2020image} for analyzing.
Please see more details and visualizations in the appendix.
As shown in the right portion of Fig.~\ref{fig:intro1}, a noteworthy observation emerges: most data points do not exhibit irregular arrangements, instead they manifest \textit{an approximately linear trend}.
% As shown in Fig.~\ref{fig:intro1-d}, surprisingly, it can be found that most 1-D datapoints do not arrange irregularly, but form an approximately increasing trend.
Among the multitude of fitting functions, the linear function stands out as the simplest yet effective one for approximating this trend.
% Among various fitting functions, linear function is the simplest one that can approximate this trend.
Inspired by this insight, we present \textbf{T}ransformer as \textbf{L}inear \textbf{E}xpansion of learn\textbf{G}ene~(\textbf{TLEG}), a novel approach for elastic Transformer production and initialization.
Specifically, we adopt linear expansion on two shared parameter modules, \textit{i}.\textit{e}., $\theta_{\mathcal{A}}$ and $\theta_{\mathcal{B}}$, both of which compose learngene $\theta_{\mathcal{LG}}$, to produce the parameters of each Transformer layer $\theta_{l}$:
% Inspired by this insight, we propose linearly expanding one shared Transformer module, \textit{i}.\textit{e}., learngene $\theta_{\mathcal{LG}}$, which contains two shared parameter tensors: $\theta_{\mathcal{A}}$ and $\theta_{\mathcal{B}}$, to produce and initialize Transformers.
\begin{equation}
\label{eq_1}
    \theta_{l}= \theta_{\mathcal{B}} + \frac{l-1}{L} \times \theta_{\mathcal{A}}, \quad l = 1,2,...,L,
\end{equation}
where $L$ denotes the total number of layers.
% Specifically, we perform such linear expansion on the parameter tensors of multi-head self-attention module, multi-layer perceptron module and layer normalization module.

\begin{figure*}[ht]
    \centering
        \includegraphics[scale=0.45]{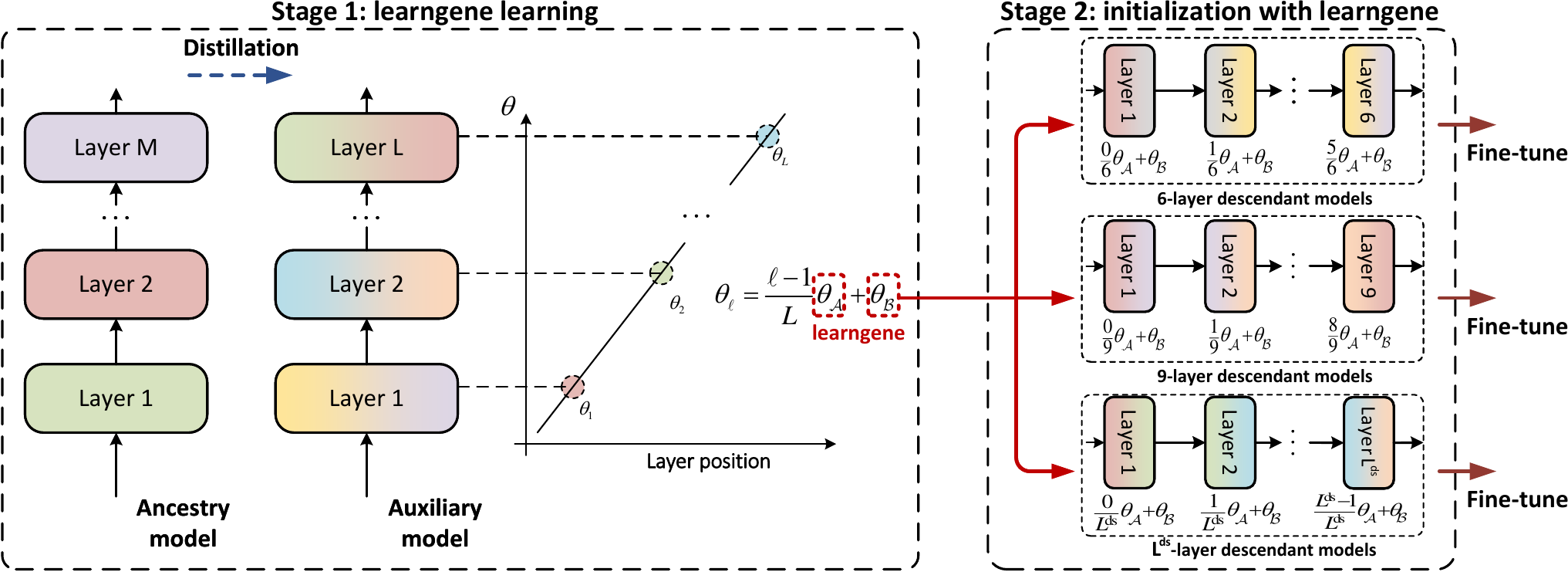}
    \caption{In the first stage, we construct an auxiliary model wherein each layer is linearly expanded from learngene.
    Subsequently, we train it through distillation.
    After obtaining learngene with well-trained $\theta_{\mathcal{A}}$ and $\theta_{\mathcal{B}}$, in the second stage, we initialize descendant models of varying depths via adopting linear expansion on $\theta_{\mathcal{A}}$ and $\theta_{\mathcal{B}}$, enabling adaptation to diverse resource constraints.
    Lastly, the descendant models are fine-tuned normally without the restriction of linear expansion.
    % In the first stage, we create an auxiliary model which is linearly expanded from learngene, after which we train it through distillation.
    % After obtaining learngene with well-trained $\theta_{\mathcal{A}}$ and $\theta_{\mathcal{B}}$, in the second stage, we initialize descendant models of varying depths via linearly expanding $\theta_{\mathcal{A}}$ and $\theta_{\mathcal{B}}$ enabling adaptation to diverse resource constraints.
    % Lastly, the descendant models are fine-tuned normally without the restriction of linear expansion.
    }
    \label{fig:method}
\end{figure*}

To learn the learngene parameters $\theta_{\mathcal{LG}}$, we design an auxiliary Transformer network (Aux-Net) where each layer is linearly expanded from $\theta_{\mathcal{A}}$ and $\theta_{\mathcal{B}}$ based on Eq.~(\ref{eq_1}).
% Given that $\theta_{\mathcal{LG}}$ is utilized to configure the parameters of each layer, we introduce patch projection, one task-oriented head and so on to construct a complete Transformer termed auxiliary model (Aux-Net) for direct training.
% Given that $\theta_{\mathcal{LG}}$ is utilized to configure the parameters of each layer, as described in Eq.~(\ref{eq_1}), we introduce patch projection and one task-oriented head to construct a complete Transformer termed auxiliary model (Aux-Net), which is amenable to direct training.
To ensure clarity, we exemplify the construction process using a 4-layer Aux-Net as an example.
The parameters of the first layer are formulated as $\theta_{1}= \theta_{\mathcal{B}} + \frac{1-1}{4} \times \theta_{\mathcal{A}}$.
Correspondingly, the parameters of the second layer are formulated as $\theta_{2}= \theta_{\mathcal{B}} + \frac{2-1}{4} \times \theta_{\mathcal{A}}$, and so forth for subsequent layers.
Then we proceed to train the Aux-Net by employing distillation technique~\cite{hinton2015distilling}, which enables knowledge condensation from a large ancestry model.
Despite the Aux-Net containing four layers, the linear constraint always holds during training, which means that only $\theta_{\mathcal{A}}$ and $\theta_{\mathcal{B}}$ will be updated throughout the training process.

After obtaining learngene containing well-trained $\theta_{\mathcal{A}}$ and $\theta_{\mathcal{B}}$, we can produce and initialize descendant models (Des-Net) of varying depths adapting for different resource constraints.
% After obtaining learngene containing well-trained $\theta_{\mathcal{A}}$ and $\theta_{\mathcal{B}}$, we can initialize the descendant models (Des-Net) with varying depths to adapt to different scenarios where diverse computation resources are available.
For example, a shallow network can be deployed in the lightweight edge device and a deeper one can be supported in a computation center equipped with ample computation resources.
To enhance clarity, we provide an example of initializing a 6-layer Des-Net. 
In this instance, the parameters of the first layer can be initialized as $\theta_{1}= \theta_{\mathcal{B}} + \frac{1-1}{6} \times \theta_{\mathcal{A}}$, similarly the parameters of the second layer would be $\theta_{2}= \theta_{\mathcal{B}} + \frac{2-1}{6} \times \theta_{\mathcal{A}}$, and so forth. 
Notably, we only employ linear expansion strategy to initialize these Des-Nets, after which they undergo standard fine-tuning procedure.
% Clearly, the transformation employed on learngene, namely linear expansion, remains coherent throughout.
Our main \textbf{contributions} are summarized as follows:
\begin{itemize}
\item We empirically discover an approximately linear relationship between the position of a layer and its corresponding weight value within well-trained Transformer models.
\item Taking inspiration from above observations, we propose a coherent approach termed TLEG for efficient model construction, which linearly expands learngene to produce and initialize Transformers across a spectrum of scales.
\item Extensive experiments demonstrate the effectiveness and efficiency of TLEG, \textit{e}.\textit{g}., compared to training different models from scratch, training with a compact learngene can obtain on-par or better performance while reducing large training costs.
\end{itemize}

%%%%%%%%%%%%%%%%%%%%%%%%%%% Related Work 
\section{Related Work}
\subsection{Parameter Initialization}
\label{RW_TL}
Parameter initialization constitutes an important step prior to model training and plays a crucial role in boosting the model quality~\cite{arpit2019initialize, huang2020improving, czyzewski2022breaking}.
Proper initialization has been proved to improve the efficiency of model training~\cite{lecun2002efficient}, whereas arbitrary initialization may impede the optimization process~\cite{mishkin2015all}.
Extensive initialization approaches have been proposed for models trained from scratch, such as random initialization, xavier initialization~\cite{glorot2010understanding}, kaiming initialization~\cite{he2016deep} and self-distillation~\cite{zhang2021self}.
% Extensive initialization schemes have been proposed, such as random initialization, xavier initialization~\cite{glorot2010understanding}, kaiming initialization~\cite{he2016deep} and self-distillation~\cite{zhang2021self}.
Nowadays, large-scale pre-training on massive curated data provides an excellent initialization for fine-tuning models across a spectrum of downstream tasks~\cite{jia2021scaling, radford2021learning, oquab2023dinov2}.
However, such scheme needs to \emph{reuse the original whole model} every time facing different downstream tasks regardless of the resources available to those tasks, as shown in Fig.~\ref{fig:intro2}(b).
% However, such scheme needs to \emph{reuse the original whole model} every time facing different downstream tasks, as shown in Fig.~\ref{fig:intro2}.
More importantly, we need to pre-train again when meets another model with different scales, which is extremely time consuming and computationally expensive.
% Moreover, we need to pre-train again when meets another model of different depths, which causes a waste of computing and storage resources.
In contrast, we propose training learngene \textit{once} which can be linearly expanded to cover a fine-grained level of model complexity/performance for a wide range of deployment scenarios.
% To alleviate the above issues, we propose to train a common and linearly expandable learngene which can be used to initialize descendant models of different depths considering both the model performance and resources, as shown in Fig.~\ref{fig:intro1-b}.

\subsection{Knowledge Distillation}
\label{RW_KD}
There exists extensive literature studying knowledge distillation~\cite{jiao2019tinybert, wang2020minilm, gou2021knowledge, wu2022tinyvit, ren2023tinymim, ji2023teachers, li2023kdcrowd}.
% There exists extensive literature studying knowledge distillation techniques~\cite{hinton2015distilling, jiao2019tinybert, wang2020minilm, gou2021knowledge, wu2022tinyvit, ren2023tinymim, ji2023teachers, li2023kdcrowd}.
% \cite{jia2021efficient} proposes building the connection between images and patches to extract knowledge from the teacher Transformer. 
DeiT~\cite{touvron2021training} introduces a distillation token to allow the vision transformer to learn from a ConvNet teacher.
MiniViT~\cite{zhang2022minivit} applies weight distillation to transfer knowledge from large-scale models to weight-multiplexed models.
TinyMIM~\cite{ren2023tinymim} studies the distillation framework for masked image modeling pretrained vision transformers.
What they have in common is that distillation requires additional forward passes through a pretrained teacher every time training a new student, which inevitably consumes extra resources for storage and computation of teacher models, as shown in Fig.~\ref{fig:intro2}(c).
In contrast, we distill rich knowledge from the pretrained ancestry model to learngene \textit{once}, after which we can produce models of diverse scales while getting rid of the ancestry model.
% What they have in common is that they need to conduct forward propagation on the teacher model every time training a new student model, which undoubtedly needs extra resources for the teacher model, as shown in Fig.~\ref{fig:intro2}.
% However, we distill knowledge from the ancestry model to learn learngene, after which we no longer need the ancestry model.

\subsection{Weight Sharing}
\label{RW_MC}
Weight sharing is a simple but effective strategy to solve over-parameterization problem~\cite{bai2019deep, kovaleva2019revealing} in large pretrained Transformers~\cite{devlin2018bert}.
By contrast, our proposed linear expansion strategy promotes parameter diversity of each layer while preserving parameter efficiency.
% we propose linearly expanding one shared module, \textit{i}.\textit{e}., learngene, to produce and initialize Transformers, such that the strategy preserves the parameter efficiency while improving the model capacity.
% By contrast, we propose to linearly expand a Transformer from a shared part, \ie, learngene, such that the operation not only preserves the parameter efficiency but also promotes the diversity.

%%%%%%%%% Approach 
\section{Approach}
\label{approach}
Fig.~\ref{fig:method} depicts the pipeline of TLEG.
In stage 1, an auxiliary model is constructed to help learn learngene parameters $\theta_{\mathcal{LG}}=\{\theta_{\mathcal{A}},\theta_{\mathcal{B}}\}$, where each layer is linearly expanded from $\theta_{\mathcal{A}}$ and $\theta_{\mathcal{B}}$.
The auxiliary model is trained by distilling knowledge from the ancestry model and note that during training, the linear constraint always holds in the auxiliary model, \textit{i}.\textit{e}., only $\theta_{\mathcal{A}}$ and $\theta_{\mathcal{B}}$ are trained.
In stage 2, the well-trained $\theta_{\mathcal{A}}$ and $\theta_{\mathcal{B}}$ are linearly expanded to initialize descendant models of varying depths.
Lastly, the descendant models are fine-tuned normally without the restriction of linear expansion.
Next, we briefly introduce some preliminaries.

% Fig.~\ref{fig:method} depicts the overall pipeline of TLEG.
% In the first stage, we construct an auxiliary model wherein each layer is linearly expanded from learngene $\theta_{\mathcal{LG}}$.
% Then we train it through distillation to condense knowledge from the large ancestry model.
% After obtaining learngene $\theta_{\mathcal{LG}}$ with well-trained $\theta_{\mathcal{A}}$ and $\theta_{\mathcal{B}}$, in the second stage, we initialize descendant models with varying depths via linearly expanding $\theta_{\mathcal{A}}$ and $\theta_{\mathcal{B}}$.
% % After obtaining learngene with well-trained $\theta_{\mathcal{A}}$ and $\theta_{\mathcal{B}}$, in the second stage, we initialize descendant models of varying depths via linearly expanding $\theta_{\mathcal{A}}$ and $\theta_{\mathcal{B}}$ for different downstream tasks.
% Lastly, the descendant models are fine-tuned normally without the restriction of linear expansion.
% % We introduce the formulation and learning of learngene in Sec.~\ref{LEL}.
% % After that, we present how to initialize the descendant model with learngene in Sec.~\ref{IL}.
% % Next, we briefly introduce some preliminaries related to the vision transformer.
% Next, we briefly introduce some preliminaries.

\subsection{Preliminaries}
Witnessing the remarkable performance of vision transformer (ViT)~\cite{dosovitskiy2020image} and its variants in diverse vision tasks~\cite{bao2021beit, wang2022image}, we explore learngene based on ViT. 
% Witnessing the remarkable performance of vision transformer (ViT)~\cite{dosovitskiy2020image} and its variants in diverse vision tasks~\cite{carion2020end, bao2021beit, wang2022image, oquab2023dinov2}, we explore learngene based on ViT. 
ViT firstly splits an image into a few patches and maps them into $D$-dimensional patch embeddings.
Then position embeddings are added to them to get $N$ input embeddings $Z_{0}\in\mathbb{R}^{N \times D}$.
A ViT encoder stacks a few layers each containing multi-head self-attention (MSA) and multi-layer perceptron (MLP).
Let $h$ denote the number of heads where each head performs self-attention.
% Let $h$ denote the number of heads where each head performs the self-attention function.
In the $k^{th}$ head, we linearly generate the queries $Q_{k}$, keys $K_{k}$ and values $V_{k}\in\mathbb{R}^{N \times d}$ with parameter matrices $W_{k}^{Q}$, $W_{k}^{K}$ and $W_{k}^{V}\in\mathbb{R}^{D \times d}$, where $d$ is the projected dimension of each head.
We denote the attention output of head $k$ as $A_k(Q, K, V) = softmax(\frac{QK^{T}}{\sqrt{d}})V$.
% \begin{equation}
% \label{eq_2}
%     A_k(Q, K, V) = softmax(\frac{QK^{T}}{\sqrt{d}})V.
% \end{equation}
MSA jointly deals with information from different embedding subspaces as $MSA(Q, K, V) = Concat(head_{1},...,head_{h})W^{O}$,
% MSA allows the model to jointly deal with information from different embedding subspaces:
% \begin{equation}
% \label{eq_3}
%     MSA(Q, K, V) = Concat(head_{1},...,head_{h})W^{O},
% \end{equation}
where $head_{k} = A_k(Q_{k}, K_{k}, V_{k})$, $W^{O}\in\mathbb{R}^{hd \times D}$ and $Concat(\cdot)$ means the catenation of the outputs of all heads.
% where $head_{k} = A_k(Q_{k}W_{k}^{Q}, K_{k}W_{k}^{K}, V_{k}W_{k}^{V})$, $W^{O}\in\mathbb{R}^{hd \times D}$ and $Concat(\cdot)$ means the catenation of the outputs of all heads.

Besides, each layer contains a MLP block which consists of two linear transformations with a GELU~\cite{hendrycks2016gaussian} activation.
We denote the MLP output as $MLP(x) = \sigma(xW^{1} + b^{1})W^{2} + b^{2}$,
% \begin{equation}
% \label{eq_4}
%     MLP(x) = \sigma(xW^{1} + b^{1})W^{2} + b^{2},
% \end{equation}
where $W^{1}\in\mathbb{R}^{D \times D_h}$, $b^{1}\in\mathbb{R}^{D_h}$, $W^{2}\in\mathbb{R}^{D_h \times D}$ and $b^{2}\in\mathbb{R}^{D}$ represent the weights and biases for the two linear transformations, respectively.
$\sigma(\cdot)$ denotes the activation function.
Usually we set $D_h$ \textgreater $D$.
Layer normalization (LN)~\cite{ba2016layer} and residual connections are employed before and after every block.
% Layer normalization (LN)~\cite{ba2016layer} and residual connections are employed before and after every block, which is crucial in Transformers.
% Layer normalization (LN)~\cite{ba2016layer} and residual connections are employed before and after every block, which is crucial in Transformer models for stable training and faster convergence.
We denote the LN output as $LN(x) = \frac{x-\mu}{\delta}\circ\gamma + \beta$,
% \begin{equation}
% \label{eq_5}
%     LN(x) = \frac{x-\mu}{\delta}\circ\gamma + \beta,
% \end{equation}
where $\mu$ and $\delta$ are the mean and standard deviation of the embeddings respectively, $\circ$ means the element-wise dot, $\gamma\in\mathbb{R}^{D}$ and $\beta\in\mathbb{R}^{D}$ are learnable transform parameters.

\subsection{Linear Expansion of Learngene}
As mentioned before, we adopt linear expansion on the two shared parameter modules, namely $\theta_{\mathcal{A}}$ and $\theta_{\mathcal{B}}$, each of which contains the parameters of an entire Transformer layer.
Since the learngene $\theta_{\mathcal{LG}}$ comprises $\theta_{\mathcal{A}}$ and $\theta_{\mathcal{B}}$, it inherently encompasses the parameters of two complete Transformer layers.
Take the 12-layer ViT-B (87M)~\cite{dosovitskiy2020image} as an example, $\theta_{\mathcal{LG}}$ comprises approximately 14.7M parameters which is equivalent to the number of parameters of two layers.
Next, we elaborate on the linear expansion of each component, \textit{i}.\textit{e}., MSA, MLP and LN, within one layer.
% As the most important part of the two stages, we firstly introduce the components of learngene and the details of linear expansion strategy below.
% As depicted in Fig.~\ref{fig:method}, in the first stage, we linearly expand learngene to form an auxiliary model.
% Then we learn the learngene through employing distillation.
% Firstly, we present the formulation and linear expansion of learngene below.

\noindent\textbf{Linear Expansion of MSA.}
Based on our empirical observations, we linearly expand the learnable parameter matrices in MSA module.
Formally, we linearly expand parameter matrices $W_{k}^{Q}$, $W_{k}^{K}$, $W_{k}^{V}$ and $W^{O}$ through Eq.~(\ref{eq_1}).
% Formally, different from the original MSA module, we linearly expand parameter matrices $W_{k}^{Q}$, $W_{k}^{K}$, $W_{k}^{V}$ and $W^{O}$ to obtain the linearly expanded version through Eq.~(\ref{eq_1}).
Take $W_{k}^{Q}$ as an example, its linearly expanded version is: 
\begin{equation}
\label{eq_6}
    W_{k}^{Q} = W_{k(\mathcal{B})}^{Q} + \frac{l-1}{L} \times W_{k(\mathcal{A})}^{Q}, \quad l = 1,2,...,L,
\end{equation}
where $W_{k(\mathcal{A})}^{Q}$ and $W_{k(\mathcal{B})}^{Q}$ are the corresponding learngene parameters of MSA in $\theta_{\mathcal{A}}$ and $\theta_{\mathcal{B}}$ respectively, $L$ denotes the total number of layers.
% where $W_{k(\mathcal{A})}^{Q}$ and $W_{k(\mathcal{B})}^{Q}$ are the corresponding learngene parameters of MSA in $\theta_{\mathcal{A}}$ and $\theta_{\mathcal{B}}$. 
% Similarly, we can generate the linearly expanded version of $W_{k}^{K}$, $W_{k}^{V}$ and $W^{O}$.
Such linear expansion can make parameters linearly different across layers while preserving the common knowledge during the training process.
% Such linear expansion can make parameters linearly different across layers while preserving the most stable and common knowledge during the training process.

\noindent\textbf{Linear Expansion of MLP.}
We further impose linear expansion on MLP to preserve the common knowledge while improving parameter diversity.
In particular, we linearly expand parameter matrices $W^{1}$, $b^{1}$, $W^{2}$ and $b^{2}$ to obtain the linearly expanded version through Eq.~(\ref{eq_1}), \textit{e}.\textit{g}., the linearly expanded version of $W^{1}$ is:
% In particular, we linearly expand parameter matrices $W^{1}$, $b^{1}$, $W^{2}$ and $b^{2}$ in Eq.~(\ref{eq_4}) to obtain the linearly expanded version through Eq.~(\ref{eq_1}), \textit{e}.\textit{g}., the linearly expanded version of $W^{1}$ is:
\begin{equation}
\label{eq_7}
    W^{1} = W^{1}_{(\mathcal{B})} + \frac{l-1}{L} \times W^{1}_{(\mathcal{A})}, \quad l = 1,2,...,L,
\end{equation}
where $W^{1}_{(\mathcal{A})}$ and $W^{1}_{(\mathcal{B})}$ are the corresponding learngene parameters of MLP in $\theta_{\mathcal{A}}$ and $\theta_{\mathcal{B}}$.
% Similarly, we can generate the linearly expanded version of $b^{1}$, $W^{2}$ and $b^{2}$.
% After the linear expansion, the outputs of MLP become more diverse, meanwhile promoting the parameter efficiency.

\noindent\textbf{Linear Expansion of LN.}
Lastly, we linearly expand learnable parameters $\gamma$ and $\beta$ through Eq.~(\ref{eq_1}).
% Lastly, we linearly expand parameters $\gamma$ and $\beta$ in Eq.~(\ref{eq_5}) to obtain the linearly expanded version through Eq.~(\ref{eq_1}).
Take $\gamma$ as an example, its linearly expanded version is:
\begin{equation}
\label{eq_8}
    \gamma = \gamma_{(\mathcal{B})} + \frac{l-1}{L} \times \gamma_{(\mathcal{A})}, \quad l = 1,2,...,L,
\end{equation}
where $\gamma_{(\mathcal{A})}$ and $\gamma_{(\mathcal{B})}$ are the corresponding learngene parameters of LN in $\theta_{\mathcal{A}}$ and $\theta_{\mathcal{B}}$.
% Similarly, we can generate the linearly expanded version of $\beta$.

\subsection{Learning Strategy of Learngene}
The learngene $\theta_{\mathcal{LG}}$ is used to construct the MSA, MLP and LN blocks by Eq.~(\ref{eq_6}) to Eq.~(\ref{eq_8}), while an integral Transformer model also requires some other components like the patch projection and task-specific head.
Thus we also add them to build the auxiliary model (\textbf{Aux-Net}), after which we train it through employing distillation. 
For simplicity, we only consider penalizing output discrepancy~\cite{hinton2015distilling} between the ancestry model and auxiliary model.
Additional distillation techniques~\cite{zhang2022minivit, ren2023tinymim} can also be seamlessly integrated into our training process, thereby further boosting the quality of the trained learngene.
Noteworthy, the linear constraint in Eq.~(\ref{eq_6}) to Eq.~(\ref{eq_8}) always exists during training.
For example, the update of $W_k^Q$ in Eq.~(\ref{eq_6}) of each layer finally leads to the update of $W_{k(\mathcal{B})}^{Q}$ and $W_{k(\mathcal{A})}^{Q}$.
Therefore, although Aux-Net contains $L$ layers, only $\theta_{\mathcal{A}}$ and $\theta_{\mathcal{B}}$ are trained during distillation.

% Since the learngene $\theta_{\mathcal{LG}}$ is employed for configuring the parameters of each layer, we further introduce additional Transformer components, such as patch projection and task-specific head, to construct a complete Transformer termed auxiliary model (Aux-Net) for direct training.
% % To learn learngene, \ie, $\theta_{\mathcal{A}}$ and $\theta_{\mathcal{B}}$, we create an auxiliary model which is linearly expanded from learngene.
% Subsequently, we proceed to train the Aux-Net through employing distillation to facilitate the training process.
% For simplicity, we only consider penalizing output discrepancy~\cite{hinton2015distilling} between the ancestry model and auxiliary model.
% Additional distillation techniques~\cite{zhang2022minivit, ren2023tinymim} can also be seamlessly integrated into our training process, thereby further boosting the quality of the obtained learngene.
% % Note that the parameters of corresponding module in auxiliary model is linearly expanded from $\theta_{\mathcal{A}}$ and $\theta_{\mathcal{B}}$ following Eq.~(\ref{Eq.6}), Eq.~(\ref{Eq.8}) and Eq.~(\ref{Eq.LN-E}).
% % For simplicity, we only consider soft distillation~\cite{hinton2015distilling}.
% Once the training process of Aux-Net finishes, the corresponding learngene parameters $\theta_{\mathcal{LG}}$ becomes proficiently trained as well.

\noindent\textbf{Soft Distillation.}
\cite{hinton2015distilling} proposes to minimize the KL-divergence between the probability distributions over their output predictions of the teacher model and that of the student one.
% \cite{hinton2015distilling} firstly proposes to minimize the Kullback-Leibler divergence between the outputs of the teacher and the outputs of the student model.
We leverage such strategy to introduce one distillation loss:
\begin{equation}
\label{Eq.9}
    \mathcal{L}_{D} = KL(\phi(z_{s}/\tau), \phi(z_{t}/\tau)),
\end{equation}
where $z_{t}$ means the logits output of the pretrained ancestry model (\textit{e}.\textit{g}., Levit-384~\cite{graham2021levit}), $z_{s}$ means the logits output of the auxiliary model, $\tau$ means the temperature for distillation, $\phi$ means the softmax function and $KL(\cdot,\cdot)$ means KL-divergence loss function.
Combined with the classification loss, our total training loss is defined as:
\begin{equation}
\label{eq_9}
    \mathcal{L} = (1-\lambda)CE(\phi(z_{s}), y) + \lambda\mathcal{L}_{D},
    % \mathcal{L} = CE(\phi(z_{s}/\tau), y) + \lambda\mathcal{L}_{D},
\end{equation}
where $y$ means ground-truth label, $CE(\cdot,\cdot)$ means cross-entropy loss function and $\lambda$ means the trade-off coefficient.

\begin{table}[t]
	\centering
	\setlength{\tabcolsep}{1.0mm}{
		\begin{tabular}{c|cccc|c|c}
			\toprule[1.5pt]
			%\cline{2-11}
			\multirow{2}*{Model} & \multirow{2}*{$D$} & \multirow{2}*{$L^{ds}$} & Params & FLOPs & \multicolumn{1}{c|}{Scratch} & \multicolumn{1}{c}{TLEG} \\
            ~ & ~ & ~ & (M) & (G) & Top-1(\%) & Top-1(\%)\\
            
            \midrule
            \multirow{4}*{Des-Ti} & \multirow{4}*{192} & 3 & 1.7 & 0.3 & 45.0 & 46.6 \textbf{(+1.6)} \\
            ~ & ~ & 6 & 3.1 & 0.6 & 56.9 & 58.2 \textbf{(+1.3)} \\
            ~ & ~ & 9 & 4.4 & 0.9 & 62.3 & 62.5 \textbf{(+0.2)} \\
            ~ & ~ & 12 & 5.7 & 1.3 & 65.2 & 65.4 \textbf{(+0.2)} \\
            \midrule
            \multirow{10}*{Des-S} & \multirow{10}*{384} & 3 & 6.1 & 1.2 & 56.2 & 57.1 \textbf{(+0.9)} \\
            ~ & ~ & 4 & 7.9 & 1.6 & 62.0 & 63.7 \textbf{(+1.7)} \\
            ~ & ~ & 5 & 9.6 & 1.9 & 67.3 & 67.5 \textbf{(+0.2)} \\
            ~ & ~ & 6 & 11.4 & 2.3 & 68.7 & 69.5 \textbf{(+0.8)} \\
            ~ & ~ & 7 & 13.2 & 2.7 & 70.6 & 71.1 \textbf{(+0.5)} \\
            ~ & ~ & 8 & 15.0 & 3.1 & 71.7 & 72.3 \textbf{(+0.6)} \\
            ~ & ~ & 9 & 16.7 & 3.5 & 73.0 & 73.2 \textbf{(+0.2)} \\
            ~ & ~ & 10 & 18.5 & 3.8 & 73.8 & 73.9 \textbf{(+0.1)} \\
            ~ & ~ & 11 & 20.3 & 4.2 & 75.5 & 75.4 (-0.1) \\
            ~ & ~ & 12 & 22.1 & 4.6 & 75.0 & 75.1 \textbf{(+0.1)} \\
            \midrule
            \multirow{10}*{Des-B} & \multirow{10}*{768} & 3 & 22.8 & 4.5 & 65.3 & 66.3 \textbf{(+1.0)} \\
            ~ & ~ & 4 & 29.9 & 5.9 & 70.4 & 71.6 \textbf{(+1.2)} \\
            ~ & ~ & 5 & 37.0 & 7.4 & 73.5 & 74.4 \textbf{(+0.9)} \\
            ~ & ~ & 6 & 44.0 & 8.8 & 75.4 & 76.2 \textbf{(+0.8)} \\
            ~ & ~ & 7 & 51.1 & 10.3 & 76.5 & 77.3 \textbf{(+0.8)} \\
            ~ & ~ & 8 & 58.2 & 11.7 & 77.2 & 78.1 \textbf{(+0.9)} \\
            ~ & ~ & 9 & 65.3 & 13.1 & 78.0 & 78.7 \textbf{(+0.7)} \\
            ~ & ~ & 10 & 72.4 & 14.6 & 78.2 & 79.1 \textbf{(+0.9)} \\
            ~ & ~ & 11 & 79.5 & 16.0 & 79.0 & 79.6 \textbf{(+0.6)} \\
            ~ & ~ & 12 & 86.6 & 17.5 & 78.6 & 79.9 \textbf{(+1.3)} \\
            
			\bottomrule[1.5pt]
		\end{tabular}
	}
	\caption{Performance comparisons on ImageNet-1K between models trained from scratch and those initialized via TLEG.}
	% \caption{Performance comparisons on ImageNet-1K between models trained from scratch with \textbf{100} epochs and those initialized via TLEG finetuned with \textbf{40} epochs.}
	\label{tab:im-1k result}
\end{table}

\subsection{Initialization with Learngene}
% \label{IL}
After obtaining learngene consisting of well-trained $\theta_{\mathcal{A}}$ and $\theta_{\mathcal{B}}$, we can produce multiple descendant models (\textbf{Des-Net}) of varying depths, catering to diverse deployment scenarios.
Benefiting from the flexibility of our proposed linear expansion strategy, we can initialize descendant models of different $L^{ds}$ by Eq.~(\ref{eq_1}).
% Benefiting from the flexibility of our proposed linear expansion strategy, we can initialize these models by linearly expanding $\theta_{\mathcal{A}}$ and $\theta_{\mathcal{B}}$:
% After learning $\theta_{\mathcal{A}}$ and $\theta_{\mathcal{B}}$ by utilizing the auxiliary model, we inherit them into different descendant models for dealing with their specific environments.
% Benefiting from the flexibility of our linear expansion strategy, we can initialize descendant models of different $L^{ds}$ with our learngene:
% \begin{equation}
% \label{eq_10}
%     \theta_{l}^{ds}= \theta_{\mathcal{B}} + \frac{l-1}{L^{ds}} \times \theta_{\mathcal{A}}, \quad l = 1,2,...,L^{ds},
% \end{equation}
% where $L^{ds}$ is the number of layers of descendant models and $\theta_{l}^{ds}$ is the parameters of $l$-th layer.
Notably, different from the Aux-Net trained under the linear constraint, the descendant models are only initialized using Eq.~(\ref{eq_6}) to Eq.~(\ref{eq_8}).
After initialization, this constraint is removed and all the parameters of the descendant models will be updated.
For example, $W_k^Q$ in Eq.~(\ref{eq_6}) of different layers will be updated normally according to their corresponding gradients irrespective of the linear constraints.

% here we linearly expand the parameters of learngene to initialize the descendant models, after which they are finetuned normally without the restriction of linear expansion.

% Notably, different from the Aux-Net trained under the restriction of linear expansion, here we linearly expand the parameters of learngene to initialize the descendant models, after which they are finetuned normally without the restriction of linear expansion.
% Note that different from the model formulation with linear expansion presented in Sec.\ref{LEL}, here we linearly expand the parameters of learngene to initialize the descendant models, after which they are trained normally without the restriction of linear expansion.
% Obviously, we only need to reuse the ancestry model \emph{once}, then we can produce and initialize descendant models with varying depths to meet diverse resource requirements of real-world deployment scenarios.
% Obviously, we only need to reuse the ancestry model \emph{once}, then we can initialize descendant models of varying depths.

%%%%%%%%%%%%%%%%%%%%%%%%%%% Experiments 
\section{Experiments}
\subsection{Experimental Setup}
% \label{es}
We conduct experiments on ImageNet-1K~\cite{deng2009imagenet} and several middle/small-scale datasets including iNaturalist 2019 (\textbf{iNat 19})~\cite{zhou2020bbn}, Mini-Imag-eNet (\textbf{Mi-INet})~\cite{vinyals2016matching}, Tiny-ImageNet (\textbf{Ti-INet})~\cite{le2015tiny}, CIFAR-10 (\textbf{C-10}), CIFAR-100 (\textbf{C-100})~\cite{krizhevsky2009learning} and Food-101 (\textbf{F-101})~\cite{bossard2014food}.
% We conduct experiments on ImageNet-1K~\cite{deng2009imagenet}, a large-scale dataset which contains about 1.2M training images and 50K validation images from 1K categories, and several middle/small-scale datasets including iNaturalist 2019 (iNat 19)~\cite{zhou2020bbn}, Mini-ImageNet (Mi-INet)~\cite{vinyals2016matching}, Tiny-ImageNet (Ti-INet)~\cite{le2015tiny}, CIFAR-10 (C-10), CIFAR-100 (C-100)~\cite{krizhevsky2009learning}, Food-101 (F-101)~\cite{bossard2014food}.
Model performance is measured by Top-1/5 accuracy (Top-1/5(\%)).
Furthermore, we report the FLOPs(G), Params(M) and S-Params(M) as indicators of theoretical complexity, the number of individual model parameters and parameters transferred/stored to initialize, respectively.
We denote Aux-Ti/S/B as the variants of Aux-Net, in which we adopt linear expansion on MSA, MLP and LN compared to DeiT-Ti/S/B~\cite{touvron2021training}.
% For brevity, we denote Aux-Ti/S/B as the auxiliary model (Aux-Net), in which we utilize linear expansion on the parameter tensors of MSA, MLP and LN compared to DeiT-Ti/S/B~\cite{touvron2021training}, thus they contain learngenes with 1.0M, 3.7M and 14.6M respectively.
For Des-Net, we introduce Des-Ti/S/B where we change the number of layers based on DeiT-Ti/S/B.
We firstly train Aux-Ti/S/B on ImageNet-1K to obtain learngenes, during which we choose Levit-384~\cite{graham2021levit} as the ancestry model to employ distillation.
% We firstly train Aux-Ti/S/B for 100 epochs on ImageNet-1K to obtain learngenes.
Then we initialize Des-Ti/S/B with learngenes and fine-tune them.
Source code is available at https://github.com/AlphaXia/TLEG.

\begin{table*}[t]
	\centering
	\setlength{\tabcolsep}{2.8mm}{
		\begin{tabular}{cc|c|ccccccc}
			\toprule[1.5pt]
			%\cline{2-11}
			\multirow{1}*{Model} & \multirow{1}*{Params(M)} & \multirow{1}*{Method} & \multirow{1}*{S-Params(M)} & \multirow{1}*{iNat 19} & \multirow{1}*{Mi-INet} & \multirow{1}*{Ti-INet} & \multirow{1}*{C-100} & \multirow{1}*{C-10} & \multirow{1}*{F-101}\\
			\midrule
            \multirow{6}*{Des-Ti} & \multirow{6}*{3.0} & \cellcolor{gray!20}Pre-Fin(U) & \cellcolor{gray!20}2.9 & \cellcolor{gray!20}58.12 & \cellcolor{gray!20}77.00 & \cellcolor{gray!20}66.32 & \cellcolor{gray!20}80.81 & \cellcolor{gray!20}96.65 & \cellcolor{gray!20}83.24 \\
            ~ & ~ & \multirow{1}*{Scratch} & 0 & 37.16 & 60.37 & 58.24 & 67.44 & 88.30 & 61.54 \\
            ~ & ~ & \multirow{1}*{He-LG} & 1.4 & 41.55 & 65.74 & 63.11 & 70.19 & 91.66 & 72.54 \\
            ~ & ~ & \multirow{1}*{Share-Init} & 0.6 & 42.58 & 63.88 & 60.98 & 71.23 & 92.56 & 67.44 \\
		    ~ & ~ & \multirow{1}*{Mini-Init} & 2.8 & 51.22 & 70.26 & 61.51 & 74.01 & 93.07 & 77.36 \\
            ~ & ~ & \multirow{1}*{TLEG(ours)} & \textbf{1.0} & \textbf{55.64} & \textbf{74.07} & \textbf{65.02} & \textbf{78.66} & \textbf{95.32} & \textbf{82.80} \\
            \midrule
            \multirow{6}*{Des-S} & \multirow{6}*{11.3} & \cellcolor{gray!20}Pre-Fin(U) & \cellcolor{gray!20}11.0 & \cellcolor{gray!20}68.48 & \cellcolor{gray!20}81.78 & \cellcolor{gray!20}72.24 & \cellcolor{gray!20}84.43 & \cellcolor{gray!20}97.59 & \cellcolor{gray!20}87.80 \\
            ~ & ~ & \multirow{1}*{Scratch} & 0 & 50.79 & 55.73 & 61.24 & 73.32 & 92.49 & 74.64 \\
            ~ & ~ & \multirow{1}*{He-LG} & 5.3 & 53.21 & 59.87 & 62.37 & 78.13 & 93.12 & 77.09\\
            ~ & ~ & \multirow{1}*{Share-Init} & 2.1 & 54.14 & 61.64 & 63.12 & 74.08 & 94.15 & 78.11\\
		    ~ & ~ & \multirow{1}*{Mini-Init} & 11.0 & 59.83 & 73.39 & 64.56 & 75.98 & 93.67 & 81.79 \\
            ~ & ~ & \multirow{1}*{TLEG(ours)} & \textbf{3.9} & \textbf{66.70} & \textbf{80.92} & \textbf{71.32} & \textbf{83.64} & \textbf{97.68} & \textbf{87.27} \\
			\bottomrule[1.5pt]
		\end{tabular}
	}
	\caption{Performance comparisons on middle/small-scale datasets when transferring pretrained parameters (S-Params(M)) to initialize 6 layer Des-Ti/S.
    Here, Params(M) means the average number of individual model parameters on different datasets.
    }
	\label{tab:result1_Des-Ti/S}
\end{table*}

\begin{table}[t]
	\centering
	\setlength{\tabcolsep}{0.6mm}{
		\begin{tabular}{c|cc|cc|cc}
			\toprule[1.5pt]
			%\cline{2-11}
			\multirow{2}*{Model}  & \multirow{2}*{$L^{ds}$} & Params & \multicolumn{2}{c|}{Pre-Fin(U)} & \multicolumn{2}{c}{TLEG} \\
            ~  & ~ & (M) & S-P(M) & Top-1(\%) & S-P(M) & Top-1(\%)\\
            
            \midrule
            \multirow{5}*{Des-B} & 4 & 29.2 & 28.8 & 87.01 & \multirow{5}*{14.7} & 86.52 \\
            % ~ & 5 & 34.6 & 35.2 &  & ~ &  \\
            ~ & 6 & 43.3 & 43.0 & 87.45 & ~ & 87.03  \\
            % ~ & 7 & 48.1 & 48.8 &  & ~ &  \\
            ~ & 8 & 57.6 & 57.1 & 88.03 & ~ & 87.96 \\
            % ~ & 9 & 61.6 & 62.3 &  & ~ &  \\
            ~ & 10 & 71.7 & 71.3 & 88.12 & ~ & 88.21 \\
            % ~ & 11 & 75.1 & 75.8 &  & ~ &  \\
            ~ & 12 & 85.9 & 85.5 & 88.62 & ~ & 88.34 \\
            
			\bottomrule[1.5pt]
		\end{tabular}
	}
	\caption{Comparisons on C-100 of Des-B with different layer numbers.
    For Pre-Fin(U), S-P(M) means the number of pretrained parameters used to initialize, which totally requires \textbf{285.7M}.
    However, TLEG only preserves \textbf{14.7M} parameters to initialize all listed Des-B, which reduces the number of parameters stored for initialization by \textbf{19$\times$ (285.7M \textit{vs}. 14.7M)}.
    }
	\label{tab:result1_Des-B_C100}
\end{table}

\begin{table}[t]
	\centering
	\setlength{\tabcolsep}{1.3mm}{
		\begin{tabular}{c|c|ccc|cc}
			\toprule[1.5pt]
			%\cline{2-11}
			\multirow{1}*{Method} & \multirow{1}*{\#} & \multirow{1}*{MSA} & \multirow{1}*{MLP} & \multirow{1}*{LN} & \multirow{1}*{Top-1 (\%)} & \multirow{1}*{Top-5 (\%)}\\
			\midrule
            \rowcolor{gray!20}
            \multirow{1}*{Pre-Fin(U)} & 1 &  &  &  & 84.43 & 96.39 \\
            \midrule
            \multirow{4}*{TLEG} & 2 & \checkmark &  &  & 79.16 & 95.17 \\
            ~ & 3 &  & \checkmark &  & 77.26 & 94.29 \\
            ~ & 4 & \checkmark & \checkmark &  & 82.03 & 95.74 \\
            ~ & 5 & \checkmark & \checkmark & \checkmark & \textbf{83.64} & \textbf{96.53} \\
			\bottomrule[1.5pt]
		\end{tabular}
	}
	\caption{Performance of 6-layer Des-S on C-100 when we employ linear expansion strategy on different modules in the 6-layer Aux-S.
    \#1 means the pre-training and fine-tuning scheme, which transfers the parameters of the total model to initialize Des-S.
    \#2/\#3/\#4/\#5 means linearly expand MSA / MLP / MSA and MLP / MSA, MLP and LN respectively.
    }
	\label{tab:ablation_1}
\end{table}

\subsection{Main Results}
\noindent\textbf{TLEG achieves on-par or better performance with much less training efforts compared to training from scratch on ImageNet-1K.}
% \noindent\textbf{TLEG achieves flexible accuracy-efficiency trade-offs.}
To validate the robustness of this claim, we conduct extensive experiments where different model settings, \textit{e}.\textit{g}., different embedding dimensions and model depths are adopted.
The ImageNet-1K classification performance of 24 different Des-Nets are reported in Table~\ref{tab:im-1k result}, where ``TLEG'' denotes the models initialized with our learngenes and ``Scratch'' denotes the randomly initialized models trained from scratch.
% The results of 18 different Des-Nets are reported in Table~\ref{tab:im-1k result}, where all models are trained on ImageNet-1K and ``TLEG'' denotes the models initialized from our learngenes and ``scratch'' denotes the randomly initialized models.
As shown in Table~\ref{tab:im-1k result}, TLEG can cover a fine-grained level of model complexity, while achieving comparable or better performance and significantly improving training efficiency.
% As shown in Table~\ref{tab:im-1k result}, TLEG can cover a wide range of accuracy-efficiency trade-offs while achieving competitive performance with an average improvement of \%.
% As it indicates, TLEG can cover a wide range of accuracy-efficiency trade-offs while achieving competitive performance with models that trained from scratch.
For Aux-S and Des-S of 10 different depths, we train Aux-S for 150 epochs and each Des-S for 35 epochs, except that we train 11-layer Des-S for 45 epochs.
From Table~\ref{tab:im-1k result}, we observe that TLEG achieves competitive performance and reduces around \textbf{2$\times$} training costs (10$\times$100 epochs \textit{vs}. 150+9$\times$35 epochs+1$\times$45 epochs), in contrast to training each Des-S from scratch for 100 epochs.
For Aux-B and Des-B of 10 different depths, we train Aux-B for 100 epochs and each Des-B for 40 epochs.
From Table~\ref{tab:im-1k result}, we find that TLEG achieves better performance and reduces around \textbf{2$\times$} training costs (10$\times$100 epochs \textit{vs}. 100+10$\times$40 epochs), compared to training each Des-B from scratch for 100 epochs.
For Aux-Ti and Des-Ti of 4 different depths, we train Aux-Ti for 150 epochs and each Des-Ti for 50 epochs.
From Table~\ref{tab:im-1k result}, we observe that TLEG achieves better performance and reduces a few training costs (4$\times$100 epochs \textit{vs}. 150+4$\times$50 epochs), contrary to training each Des-Ti from scratch for 100 epochs.
Overall, the efficiency of TLEG becomes more evident with the number of Des-Nets increasing as we only need to train our learngenes \emph{once}.
% Also we notice that the efficiency of TLEG becomes more evident with the number of Des-Nets increasing as we only need to train our learngenes \emph{once}.
% Take Des-B of 10 different depths as an example, TLEG achieves competitive performance and reduces around \textbf{2$\times$} training costs (10$\times$100 epochs \textit{vs}. 100+10$\times$40 epochs), in contrast to training individual Des-Nets from scratch.
% Take Des-B of 10 different depths as an example, TLEG reduces the training cost by around \textbf{2$\times$} (10$\times$100 epochs \textit{vs}. 100+10$\times$40 epochs) and the number of parameters used for initialization by \textbf{37$\times$} (521.4M \textit{vs}. 14.0M), in contrast to training and initializing individual Des-Nets.
% To be emphasized, TLEG reduces around 2$\times$ training cost (18 $\times$ 120 epochs \textit{vs}. 3 $\times$ 100 + 18 $\times$ 40 epochs) and 31$\times$ local disk storage (589.3M \textit{vs}. 18.7M) compared to training and saving individual Des-Nets.

% Furthermore, the efficiency of TLEG becomes more evident with the number of Des-Nets increasing.
% Also we notice that when Params(M) is tiny, TLEG performs slightly worse due to less discrepancy between parameters randomly initialized and those initialized via TLEG, but TLEG still maintains training efficiency.

\noindent\textbf{TLEG provides competitive results when transferring to a wide range of downstream classification datasets.}
% \noindent\textbf{TLEG provides competitive results when downstream model is consistent with the pretrained model.}
We compare TLEG against training from scratch and pre-training method whose performance is regarded as upper-bound on 6 classification datasets.
% We compare TLEG against random initialization and popular pre-training method whose performance is regarded as upper-bound.
Moreover, we adopt state-of-the-art compression methods to our setting. 
Specifically,
(1)~Scratch.
We train Des-Nets from scratch on the downstream datasets.
% We initialize Des-DeiT randomly and then training it on the downstream dataset.
(2)~Pre-Fin(U).
We pretrain each Des-Net on ImageNet-1K with 100 epochs.
% We pretrain each Des-Net on ImageNet-1K with 100 epochs to keep fair comparison.
% We pre-train Des-DeiT on the data which we use to train the learngene, \textit{i}.\textit{e}., ImageNet-1K, to keep fair comparison.
% Then we fine-tune whole model on the downstream dataset.
(3)~Mini-Init~\cite{zhang2022minivit}.
We pretrain Mini-DeiT on ImageNet-1K with 100 epochs, where the number of shared part is 6.
Then we use the shared parts to initialize Des-Nets.
% We pre-train Mini-DeiT on ImageNet-1K, which share the parameters every two consecutive layers.
% As Mini-DeiT stacks 12 layers, the number of shared part is 6.
% Then we use the 6 shared parts to initialize Des-DeiT.
(4)~Share-Init~\cite{lan2019albert}.
We pre-train DeiT~\cite{touvron2021training} on ImageNet-1K with 100 epochs, where we share the parameters of each layer.
Then we use the shared part to initialize Des-Nets.
% As weight sharing causes unstable training process and less model capacity shown in~\cite{zhang2022minivit}, we pre-train DeiT-Ti and DeiT-S~\cite{touvron2021training} on ImageNet-1K following~\cite{zhang2022minivit}, where we share the parameters of each layer. 
% Then we use the shared part to initialize each layer of Des-DeiT.
(5)~He-LG~\cite{wang2022learngene}.
We extract last 3 layers of pretrained DeiTs and stack them with randomly-initialized low-level layers to produce Des-Nets.
% We pre-train DeiT-Ti and DeiT-S~\cite{touvron2021training} on ImageNet-1K.
% Then we extract last 3 layers and stack them with randomly-initialized low-level layers to initialize Des-DeiT.
For (2)-(5), we finetune Des-Nets on the downstream datasets.
As shown in Table~\ref{tab:result1_Des-Ti/S}, TLEG achieves performance gains compared with several baselines, which verifies the effectiveness of initialization with learngenes.
% As shown in Table~\ref{tab:result1_Des-Ti/S}, TLEG achieved significant performance gains on all downstream datasets compared with several baselines, which verifies the effectiveness of initialization with learngenes.
For example, we observe that TLEG outperforms Mini-Init by \boldmath{$7.53\%$, $6.76\%$, and $7.66\%$} respectively on Mi-INet, Ti-INet and C-100 with Des-S, whereas TLEG reduces \textbf{2.8$\times$} parameters used to initialize (11.0M \textit{vs}. 3.9M).
% only need to store and transfer about \textit{a third of} the total parameters to the Des-S.
Moreover, TLEG achieves comparable performance compared with upper-bound method Pre-Fin(U), showing that the common knowledge, \textit{i}.\textit{e}., learngene, is satisfactorily learned and used to initialize Des-Nets.

\noindent\textbf{TLEG significantly reduces the parameters stored to initialize and pre-training costs compared with Pre-Fin(U) when initializing diverse models.}
We compare 5 different Des-B initialized from learngenes to those initialized via Pre-Fin(U), where the performance of latter is regarded as upper-bound.
In Table~\ref{tab:result1_Des-B_C100}, TLEG achieves comparable performance and efficiently initializes diverse models with fewer storage costs.
Specifically, TLEG substantially reduces \textbf{19$\times$} (285.7M \textit{vs}. 14.7M) parameters stored to initialize, compared to Pre-Fin(U).
Moreover, Pre-Fin(U) needs to pretrain each different Des-B individually, while TLEG only requires training learngene \textit{once}, thus substantially reducing the pre-training costs.
Specifically, TLEG reduces \textbf{5$\times$} (5$\times$100 epochs \textit{vs}. 1$\times$100 epochs) pre-training costs compared to Pre-Fin(U) when facing 5 different Des-Nets.
Notably, the efficiency of TLEG becomes more obvious because the pre-training costs of Pre-Fin(U) increase with the number of different Des-Nets.

\begin{table}[t]
	\centering
	\setlength{\tabcolsep}{4.0mm}{
		\begin{tabular}{c|c|cc}
			\toprule[1.5pt]
			%\cline{2-11}
			\multirow{1}*{Model} & \multirow{1}*{Part} & \multirow{1}*{Top-1 (\%)} & \multirow{1}*{Top-5 (\%)}\\
			\midrule
            \multirow{3}*{Des-Ti} & 1 - 3 & 74.29 & 93.58 \\
		    ~ & 4 - 6  & 72.00  & 92.55 \\
            ~ & 1 - 6  & \textbf{78.66}  & \textbf{95.35} \\
            \midrule
            \multirow{3}*{Des-S} & 1 - 3 & 74.88 & 93.59 \\
		    ~ & 4 - 6 & 74.33 & 93.33 \\
            ~ & 1 - 6 & \textbf{83.64} & \textbf{96.53} \\
			\bottomrule[1.5pt]
		\end{tabular}
	}
	\caption{Performance of 6-layer Des-Ti/S on C-100 under partially initialization.
    ``Part'' means the scope of the layers we initialize, \textit{e}.\textit{g}., ``1 - 3'' means we initialize the first 3 layers.
    }
	\label{tab:ablation_2}
\end{table}

\begin{figure*}[ht]
    \centering
    \begin{subfigure}{0.33\linewidth}
        \includegraphics[scale=0.24]{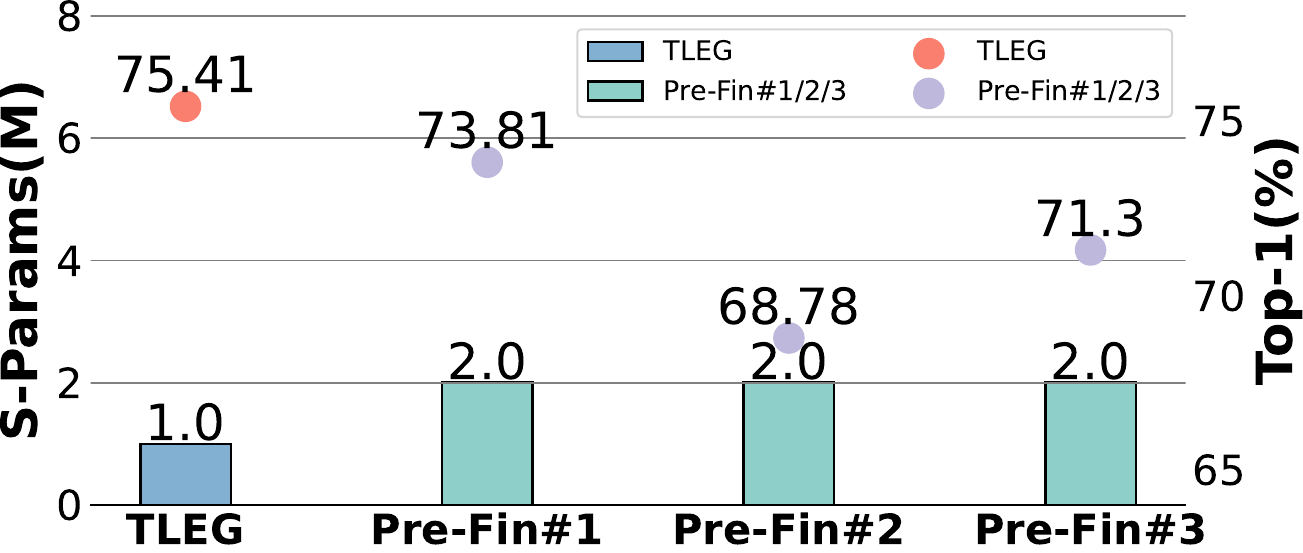}
        \caption{4-layer Des-Ti}
        \label{fig:Ti-1}
    \end{subfigure}
    \hfill
    \begin{subfigure}{0.33\linewidth}
        \includegraphics[scale=0.24]{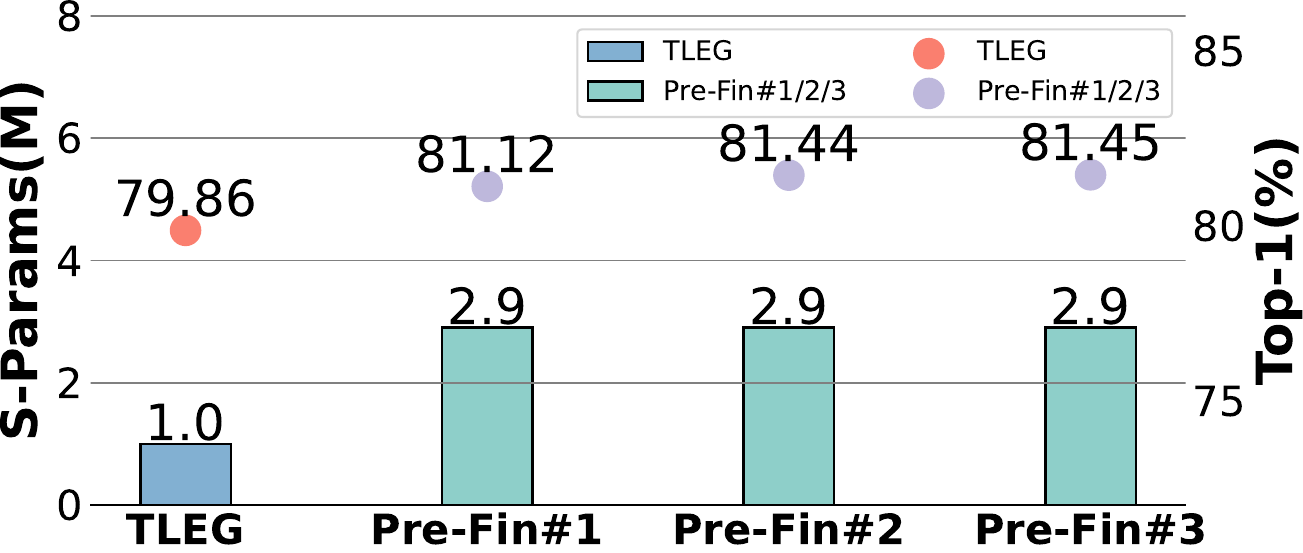}
        \caption{8-layer Des-Ti}
        \label{fig:Ti-2}
    \end{subfigure}
    \hfill
    \begin{subfigure}{0.33\linewidth}
        \includegraphics[scale=0.24]{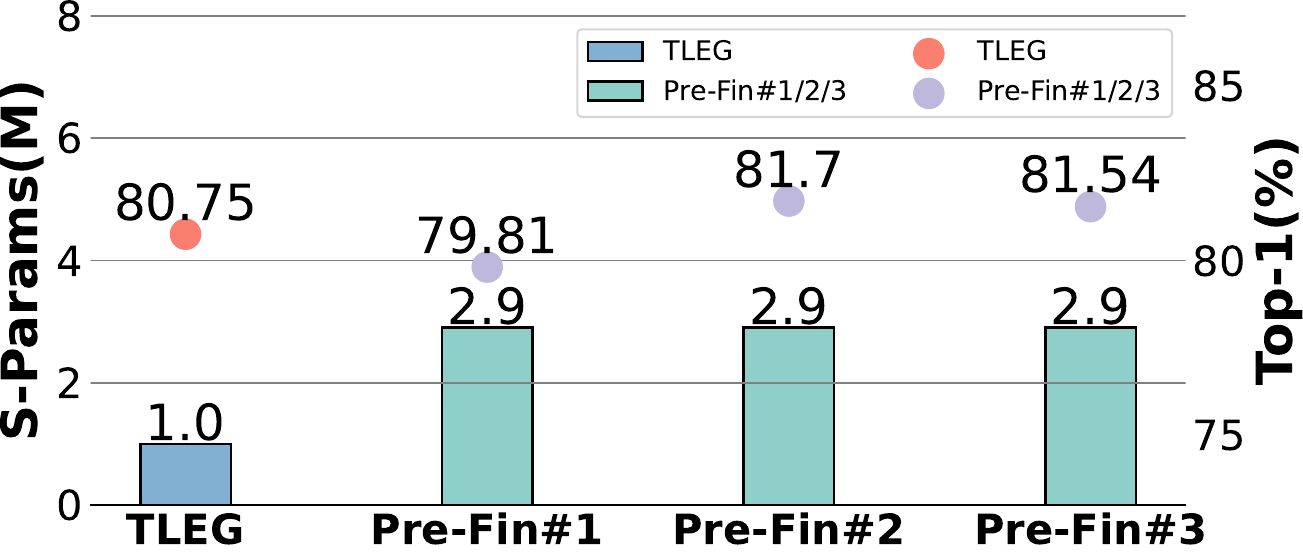}
        \caption{12-layer Des-Ti}
        \label{fig:Ti-3}
    \end{subfigure}
    
    \begin{subfigure}{0.33\linewidth}
        \includegraphics[scale=0.24]{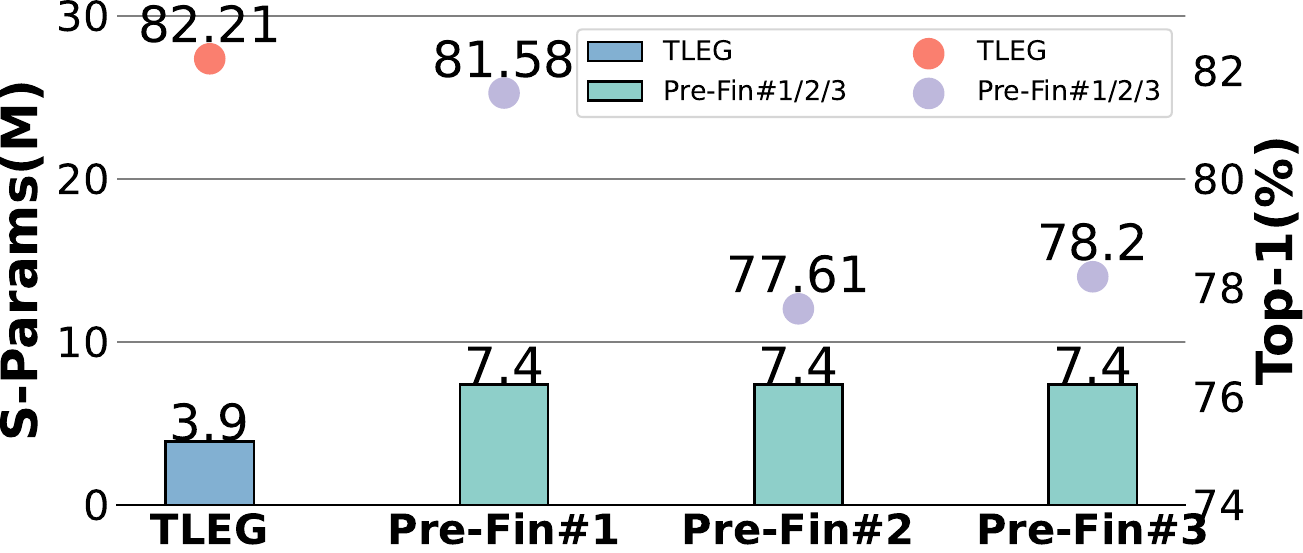}
        \caption{4-layer Des-S}
        \label{fig:s-1}
    \end{subfigure}
    \hfill
    \begin{subfigure}{0.33\linewidth}
        \includegraphics[scale=0.24]{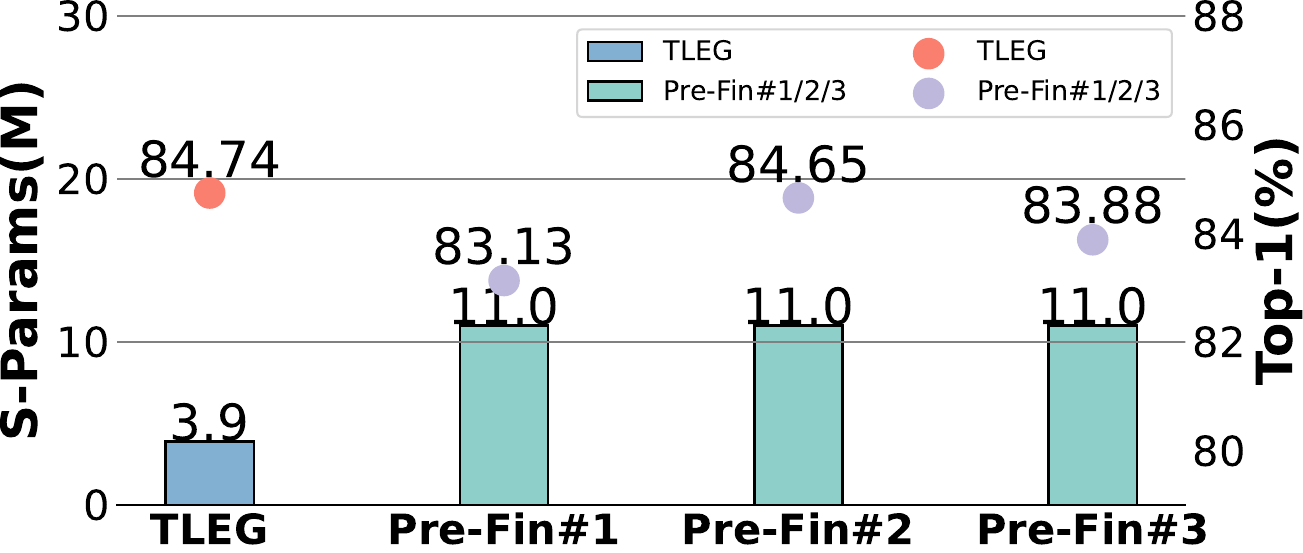}
        \caption{8-layer Des-S}
        \label{fig:s-2}
    \end{subfigure}
    \hfill
    \begin{subfigure}{0.33\linewidth}
        \includegraphics[scale=0.24]{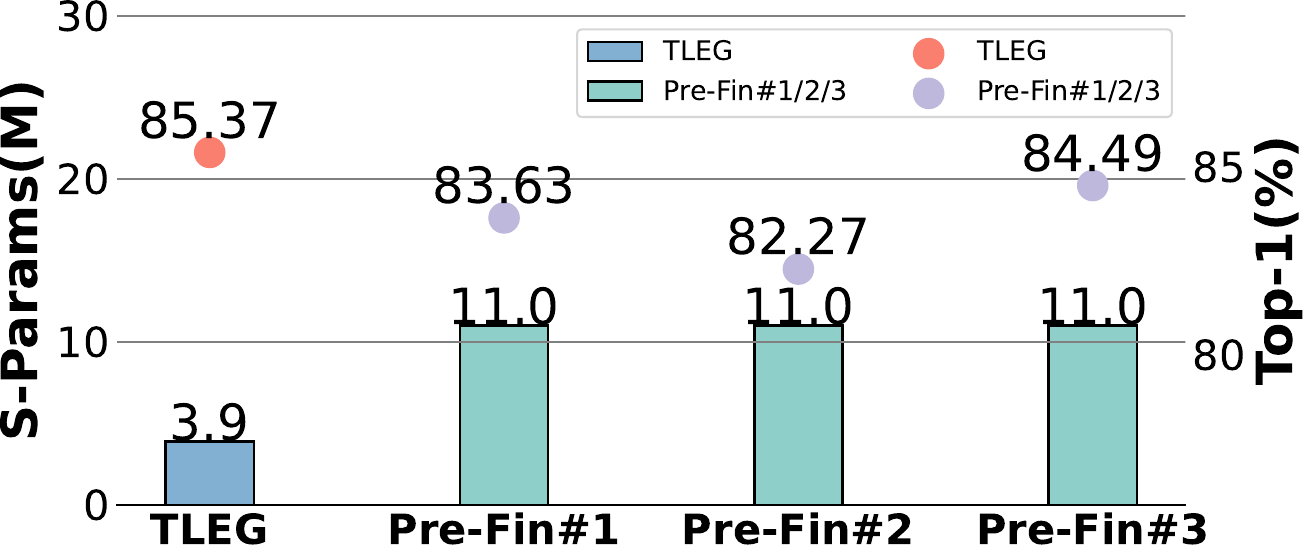}
        \caption{12-layer Des-S}
        \label{fig:s-3}
    \end{subfigure}

    \begin{subfigure}{0.33\linewidth}
        \includegraphics[scale=0.24]{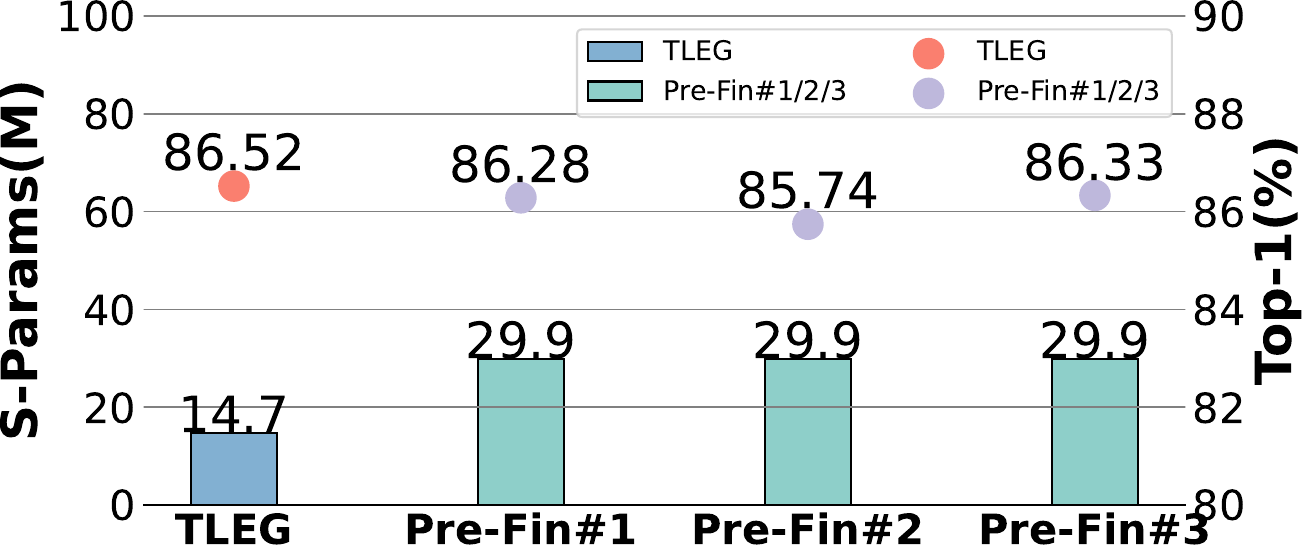}
        \caption{4-layer Des-B}
        \label{fig:b-1}
    \end{subfigure}
    \hfill
    \begin{subfigure}{0.33\linewidth}
        \includegraphics[scale=0.24]{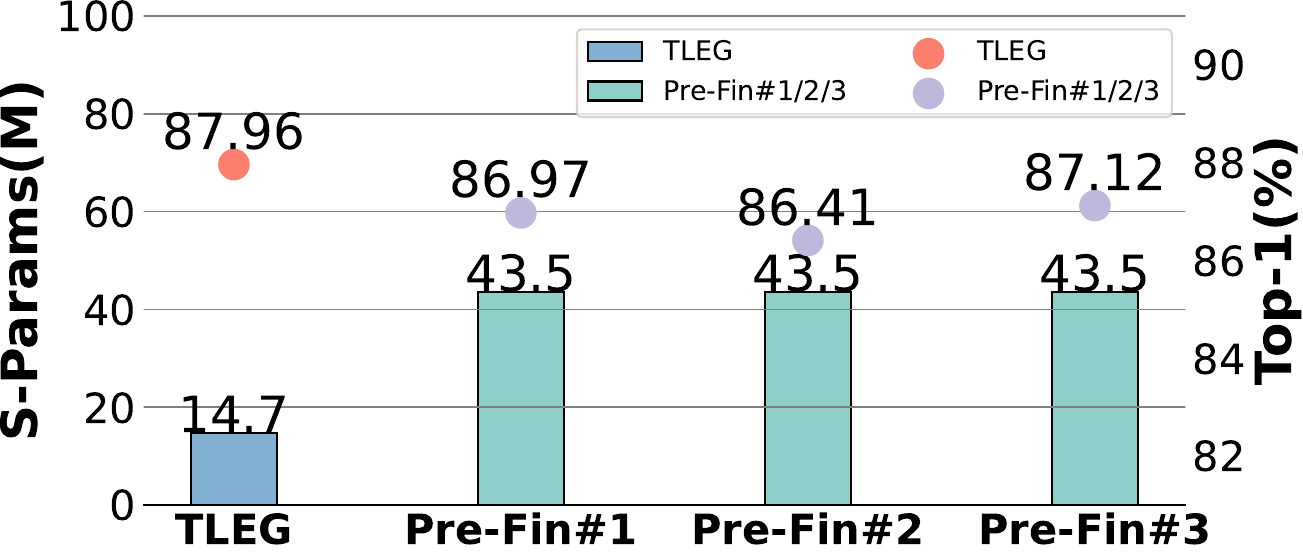}
        \caption{8-layer Des-B}
        \label{fig:b-2}
    \end{subfigure}
    \hfill
    \begin{subfigure}{0.33\linewidth}
        \includegraphics[scale=0.24]{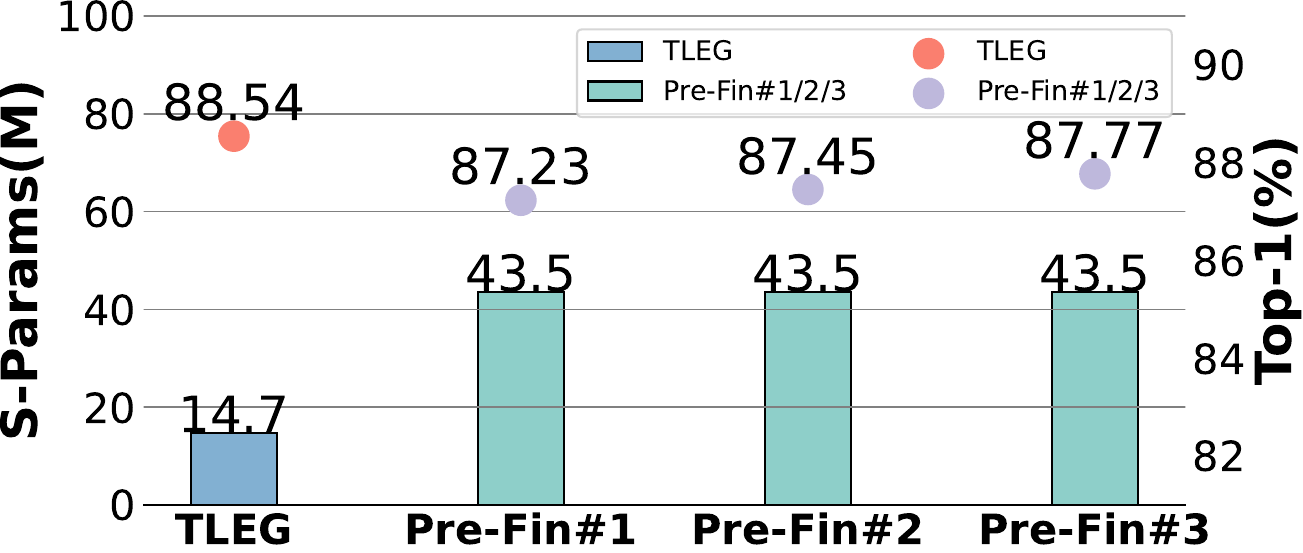}
        \caption{12-layer Des-B}
        \label{fig:b-3}
    \end{subfigure}
    
    \caption{Performance on C-100 when initializing diverse Des-Nets of different scales with a fixed set of parameters.
    % pretrained model (6-layer Aux-Ti/S/B) is inconsistent with downstream model.
    % \caption{Performance comparsion on C-100 of Des-Ti/S/B when pretrained model (6-layer Aux-Ti/S/B) is inconsistent with downstream model (a)/(d)/(g) 4-layer Des-Ti/S/B, (b)/(e)/(h) 8-layer Des-Ti/S/B and (c)/(f)/(i) 12-layer Des-Ti/S/B.
    }
    \label{fig:result2_Des-Ti/S}
\end{figure*}

\noindent\textbf{TLEG presents better performance and flexibility when initializing models of different scales with a fixed set of parameters.}
% \noindent\textbf{TLEG presents better performance and flexibility when initializing various models of different scales with a fixed/constrained set of parameters.}
% \noindent\textbf{TLEG presents better flexibility and performance when downstream model is \emph{inconsistent} with the pretrained model.}
Specifically, we have three 6-layer model of different embedding dimensions pretrained on Imagenet-1K via Pre-Fin(U) with 57.4M (2.9+11.0+43.5M) parameters and relatively smaller learngenes with 19.6M (1.0+3.9+14.7M) parameters.
Now we need to initialize the 4-layer, 8-layer and 12-layer Des-Ti/S/B.
For TLEG, we initialize them via linearly expanding the learned learngenes conveniently.
For Pre-Fin(U), we have several intuitive choices:
For the 8-layer and 12-layer Des-Ti/S/B, Pre-Fin \#1/\#2/\#3 means we initialize the first/last/middle 6 layer of them with the pretrained 6-layer models.
For the 4-layer Des-Ti/S/B, Pre-Fin \#1/\#2/\#3 means that we use the first/last/middle 4 layer of 6-layer pretrained models to initialize them.
As shown in Fig.~\ref{fig:result2_Des-Ti/S}, we observe that TLEG achieves comparable or even superior performance over Pre-Fin(U) while reducing around \textbf{2.9$\times$} (57.4M \textit{vs}. 19.6M) parameters stored to initialize.
For example, TLEG outperforms the best variant of Pre-Fin(U) by \boldmath{$0.63\%$, $0.09\%$ and $0.88\%$} respectively on 4-layer, 8-layer and 12-layer Des-S.
Overall, when initializing different models with a fixed set of parameters, TLEG demonstrates better flexibility and performance, showing that learngene contains generalizable knowledge and serves as a great starting point for training Des-Nets of diverse scales.

% Overall, when initializing different models with a fixed/constrained set of parameters, TLEG demonstrates better flexibility and performance, showing that learngene contains generalizable knowledge and serves as a great starting point for training Des-Nets of diverse scales.
% Overall, when initializing a wide range of models with different scales, TLEG demonstrates better flexibility and performance, showing that learngene contains generalizable knowledge and serves as a great starting point for training Des-Nets of diverse scales.
% Such experimental results show the effectiveness and flexibility of TLEG.
% Such experimental results show the effectiveness and efficiency of TLEG for flexible initialization when downstream model is inconsistent with the pre-trained model.

\begin{table}[t]
	\centering
	\setlength{\tabcolsep}{4.0mm}{
		\begin{tabular}{c|c|cc}
			\toprule[1.5pt]
			%\cline{2-11}
			\multirow{1}*{Model} & \multirow{1}*{Type} & \multirow{1}*{Top-1 (\%)} & \multirow{1}*{Top-5 (\%)}\\
			\midrule
            \multirow{2}*{Des-Ti} & All & 78.66 & 95.35 \\
            ~ & Partial  & \textbf{79.02}  & \textbf{96.12} \\
            \midrule
            \multirow{2}*{Des-S} & All & 83.64 & 96.53 \\
            ~ & Partial & \textbf{84.13} & \textbf{96.72} \\
			\bottomrule[1.5pt]
		\end{tabular}
	}
	\caption{Performance of 6-layer Des-Ti/S on C-100 with different linear
    expansion strategies.
    ``Partial'' and ``All'' mean that we adopt linear expansion from the 3rd layer or on all layers in 6-layer Aux-Ti/S, respectively.
    }
	\label{tab:ablation_3}
\end{table}

\subsection{Ablation and Analysis}
We investigate the performance of Des-Nets when we (1) adopt linear expansion on different modules in Aux-Nets, (2) initialize partial layers of Des-Nets, (3) adopt linear expansion on partial layers in Aux-Nets.

% In this subsection, we investigate the performance of Des-Nets when we (1) adopt linear expansion on partial layers in Aux-Nets, (2) adopt linear expansion on different modules in Aux-Nets, (3) initialize partial layers of Des-Nets.
% We provide more ablation studies in the appendix.
% \label{aa}
% In this subsection, we investigate the performance of Des-Nets when we (1) adopt linear expansion on different modules in auxiliary model, (2) change the depth of descendant models and (3) initialize partial layers of descendant models.
% We report the results on CIFAR-100.

\noindent\textbf{The effect of different linearly expanded modules.}
We apply linear expansion strategy on different modules of Aux-S to achieve several variants.
Then we utilize the linearly expanded module to initialize corresponding module in Des-S and randomly initialize other modules.
% For example, we linearly expand the MSA module in Aux-S.
% After training it, we use the trained MSA module to initialize the MSA module in Des-S with linear expansion strategy and randomly initialize other modules.
% Lastly, we fine-tune Des-S.
As shown in Table~\ref{tab:ablation_1}, we can observe that \#5 achieves the best accuracy, which is comparable against Pre-Fin(U).
% Furthermore, \#5 reduces around \textbf{2.8$\times$} (11.0M \textit{vs}. 3.9M) parameters stored to initialize.
% needs to transfer learngene which is about a third of total model parameters. 

\noindent\textbf{The effect of initializing partial layers of Des-Nets.}
We initialize partial layers of 6-layer Des-S.
As shown in Table~\ref{tab:ablation_2}, we choose to initialize first half~(1-3), second half~(4-6) and all~(1-6) of the layers.
We observe that initializing all layers achieves the best performance.
% \paragraph{Initialization part of descendant model.}
% We investigate the performance of Des-DeiT-S when we initialize partial layers.
% As shown in Table~\ref{tab:ablation_3}, we choose to initialize first half of the layers~(1-3), second half of the layers~(4-6), and all layers~(1-6).
% We observe that initializing all layers achieves the best performance, which verifies the effectiveness of our learned learngene.

\noindent\textbf{The effect of partial linear expansion.}
We adopt linear expansion on partial layers, \textit{i}.\textit{e}., from the 3rd layer to the last layer, in 6-layer Aux-Ti/S to learn learngene.
As shown in Table~\ref{tab:ablation_3}, we observe that adopting partial linear expansion achieves slightly better performance.
Nevertheless, we adopt linear expansion on all layers in our main experiments.
More variants of linear expansion strategies are left for future work.

%%%%%%%%%%%%%%%%%%%%%%%%%%% Conclusion 
\section{Conclusion}
\label{sec:conclusion}
In this paper, we proposed a new approach termed TLEG to produce and initialize Transformers of varying depths via linearly expanding learngene, enabling adaptation to diverse real-world applications containing different resources.
Experimental results under various model initialization settings demonstrated the effectiveness and flexibility of TLEG.

%%%%%%%%%%%%%%%%%%%%%%%%% Acknowledgments %%%%%%%%%%%%%%%%%%%%%%%%%
\section*{Acknowledgements}
This research is supported by the National Key Research \& Development Plan of China (No. 2018AAA0100104), the National Science Foundation of China (62125602, 62076063), National Science Foundation of China (62206048), Natural Science Foundation of Jiangsu Province (BK20220819), Young Elite Scientists Sponsorship Program of Jiangsu Association for Science and Technology Tj-2022-027, and the Big Data Computing Center of Southeast University.

\bibliography{aaai24}

\end{document}